\documentclass[10pt, a4paper, onecolumn]{article}

\usepackage[letterpaper,top=2cm,bottom=2cm,left=3cm,right=3cm,marginparwidth=1.75cm]{geometry}

\usepackage[numbers]{natbib}
\usepackage[english]{babel}
\usepackage[T1]{fontenc}
\usepackage{lmodern}
\usepackage{natbib}
\usepackage[utf8]{inputenc} 
\usepackage{nicefrac}       
\usepackage{microtype}      
\usepackage{bm}
\usepackage{url}

\PassOptionsToPackage{table}{xcolor}
\usepackage{graphicx}
\usepackage{multirow}
\usepackage{caption}
\usepackage{subcaption}
\usepackage{mathtools}
\usepackage{booktabs}
\usepackage[table]{xcolor}  
\usepackage[squaren]{SIunits}
\usepackage{array}
\usepackage{balance}
\usepackage{hyperref}
\newcommand{\figref}[1]{Fig.~\ref{#1}}
\newcommand{\subfigref}[1]{(\subref{#1})}
\newcommand{\tblref}[1]{Table~\ref{#1}}

\newcommand{\pluseq}{\mathrel{+{=}}}
\newcommand{\round}[1]{\ensuremath\left\lfloor#1\right\rceil}

\newcommand{\pz}{\phantom{0}}

\title{Gaze-based Object Detection in the Wild}

\author{
	Daniel Weber\thanks{Both authors contributed equally to this research.} \thanks{Funded by the Deutsche Forschungsgemeinschaft (DFG, German Research Foundation) under Germany’s Excellence Strategy -- EXC number 2064/1 -- Project number 390727645.} \\ University of Tübingen \\ \texttt{daniel.weber@uni-tuebingen.de}
	\and
	Wolfgang Fuhl\footnotemark[1] \\ University of Tübingen \\ \texttt{wolfgang.fuhl@uni-tuebingen.de}
	\and
	Andreas Zell \\ University of Tübingen \\ \texttt{andreas.zell@uni-tuebingen.de}
	\and
	Enkelejda Kasneci \\ Technical University of Munich \\ \texttt{enkelejda.kasneci@tum.de}
}

\begin{document}

	\maketitle
	
	\begin{abstract}
		In human-robot collaboration, one challenging task is to teach a robot new yet unknown objects enabling it to interact with them.
Thereby, gaze can contain valuable information.
We investigate if it is possible to detect objects (object or no object) merely from gaze data and determine their bounding box parameters.
For this purpose, we explore different sizes of temporal windows, which serve as a basis for the computation of heatmaps, i.e., the spatial distribution of the gaze data.
Additionally, we analyze different grid sizes of these heatmaps, and demonstrate the functionality in a proof of concept using different machine learning techniques.
Our method is characterized by its speed and resource efficiency compared to conventional object detectors.
In order to generate the required data, we conducted a study with five subjects who could move freely and thus, turn towards arbitrary objects.
This way, we chose a scenario for our data collection that is as realistic as possible.
Since the subjects move while facing objects, the heatmaps also contain gaze data trajectories, complicating the detection and parameter regression.
We make our data set publicly available to the research community for download.
	\end{abstract}	
	
	\section{Introduction}
Recent research has shown that eye tracking has becoming increasingly relevant for a variety of applications. These include even dynamic real-world scenarios, such as driving \cite{palinko2010estimating}, \cite{braunagel2016necessity}, \cite{xu2018real}, medicine \cite{van2017visual}, \cite{harezlak2018application}, \cite{tien2010measuring}, and sports \cite{grushko2014usage}, \cite{hosp2021soccer}, \cite{panchuk2015eye}.
Especially the combination with computer vision problems \cite{toyama2012gaze}, \cite{shanmuga2015eye}, has in turn great potential for the employment of eye tracking in other fields, such as robotics \cite{palinko2016robot},\cite{chadalavada2020bi}, \cite{weber2020distilling}.
In the field of robotics, the focus is often on the interaction with the environment, for example, detecting and grasping objects \cite{kehoe2013cloud}, \cite{mahler2019learning}.
In such settings, however, the interaction entities are often unknown due to the enormous amount of potentially existing objects.
For this purpose, a semantic understanding of scenes must be present.
In conveying this understanding, humans can play an important role and provide assistance to the robot.
One modality that has proven to be particularly suitable and helpful for such human-robot collaboration settings is the human gaze \cite{weber2022exploiting}.
Gaze allows objects to be intuitively selected by the human and communicated (e.g., gaze pointing) to the interaction partner (e.g., robot). 
An additional advantage of the gaze modality is that it is far more unambiguous than gestures and, unlike speech, can also be used effortlessly in the case of unknown objects whose class name may not be known at all.

In this work, we address the problem of unknown object detection in real-world scenarios based on gaze.
This is an essential challenge for human-robot collaboration, as an example.
After all, if the robot could detect an unknown object by the fact that the human is looking at it, this paves the way for further interaction possibilities.
We refer to object detection in a similar manner to face detection.
In face detection, the task is to estimate whether there is a face or not.
In our task, the challenge is to find out whether the current gaze pattern belongs to a perceived object or not.
While there is work investigating unknown object detection on static imagery, there is little research addressing unknown object detection on videos and settings in the wild.
Along this line, \cite{li2014secrets} and \cite{xiao2018salient} used fixations to infer the saliency of objects.
A gaze map was used by \cite{shi2017gaze}, who combined it with candidate regions to segment objects. In addition, the authors in  \cite{yun2013exploring} investigated the relationship between fixations made while observing an image and the object categories it contains.
The finding was that machine learning models can benefit from human fixations for detection and classification tasks.
In the work by \cite{luo2019interested}, gaze points were grouped into clusters to determine whether a cluster belonged to an object of interest and whether it was looked at intentionally or unintentionally.
However, all these related works used multiple gaze points on one image, which is only possible if the stimulus (image of the observed scene) is static or if, for instance, eye tracking data from multiple people is used, as in \cite{shanmuga2015eye}.
In the latter, it was possible to extract attentionally important objects from videos.
Contrary to all aforementioned related works, we present a method capable of using gaze data from a single person in dynamic scenes, i.e., with non-static stimuli, to detect unknown objects.

Our way to meet this challenge is by considering and analyzing gaze data across multiple frames and constructing a heatmap from it. 
In contrast, \cite{weber2020distilling} significantly reduced the amount of candidate bounding boxes of unknown objects on a static image using only one gaze point.
In another recent work in a human-robot collaboration scenario, \cite{weber2022exploiting} achieved segmentation of unknown objects and calculated corresponding bounding boxes in 3D space in real time.
Although only one gaze point was required here, the scene image including depth information was needed.
Some other approaches dispense with the gaze altogether, but focus rather on single-class images \cite{pang2020multi}, \cite{kootstra2010fast}, or use additional information, e.g., from a depth sensor \cite{bao2015saliency}.
While robots typically have many sensors, they often have limited computing power and, in some cases, not even a graphics processor, so using large amounts of image data, especially depth images, puts a lot of strain on them.
Additionally, there is often only one object of interest at a time, obviating the need to detect all objects at once. 
By completely omitting image data and employing gaze data instead, we can accomplish the task of detecting unknown objects of interest and still saving large amounts of required computer resources.

In this work, we build on existing work and pave the way for successful human-robot interaction through the following  main contributions:
\begin{itemize}
    \item We present a method for detecting unknown objects in a scene without stimulus, based solely on gaze information. 
    \item We only use heatmaps instead of scene images, enabling thus for a significantly faster approach than image-based object detection, while at the same time requiring considerably less computational resources.
    \item In a proof of concept, we evaluate the accuracy of different machine learning techniques and parameters, such as the time window size for the spatial distribution of the gaze data and the grid size of the resulting heatmaps.
    \item We make our unique data set, which contains both gaze data and bounding boxes of the observed objects, publicly available to the research community for download at \url{https://cloud.cs.uni-tuebingen.de/index.php/s/QPzJC48xDGsjnZK}.
\end{itemize}
	\section{Method}
In this work, we challenge object detection by means of gaze points. Here we follow two goals. First, we classify which gaze points or ranges of gaze points belong to an object, and we assign temporal windows to the gaze points, which belong to an annotated bounding box. This creates a classification problem in which the gaze points windows with an associated bounding box are assigned to class one and gaze points windows without a bounding box are assigned to class zero. 

The second goal is to regress the bounding box parameters on the gaze points. These parameters are the width and height, as well as the x and y position. For this task, we also assigned the gaze points to temporal windows. For the regression, we used only temporal windows with associated bounding box, since all other temporal windows have no parameters for the regression.

\begin{figure}[htb]
	\centering
	\includegraphics[width=\linewidth]{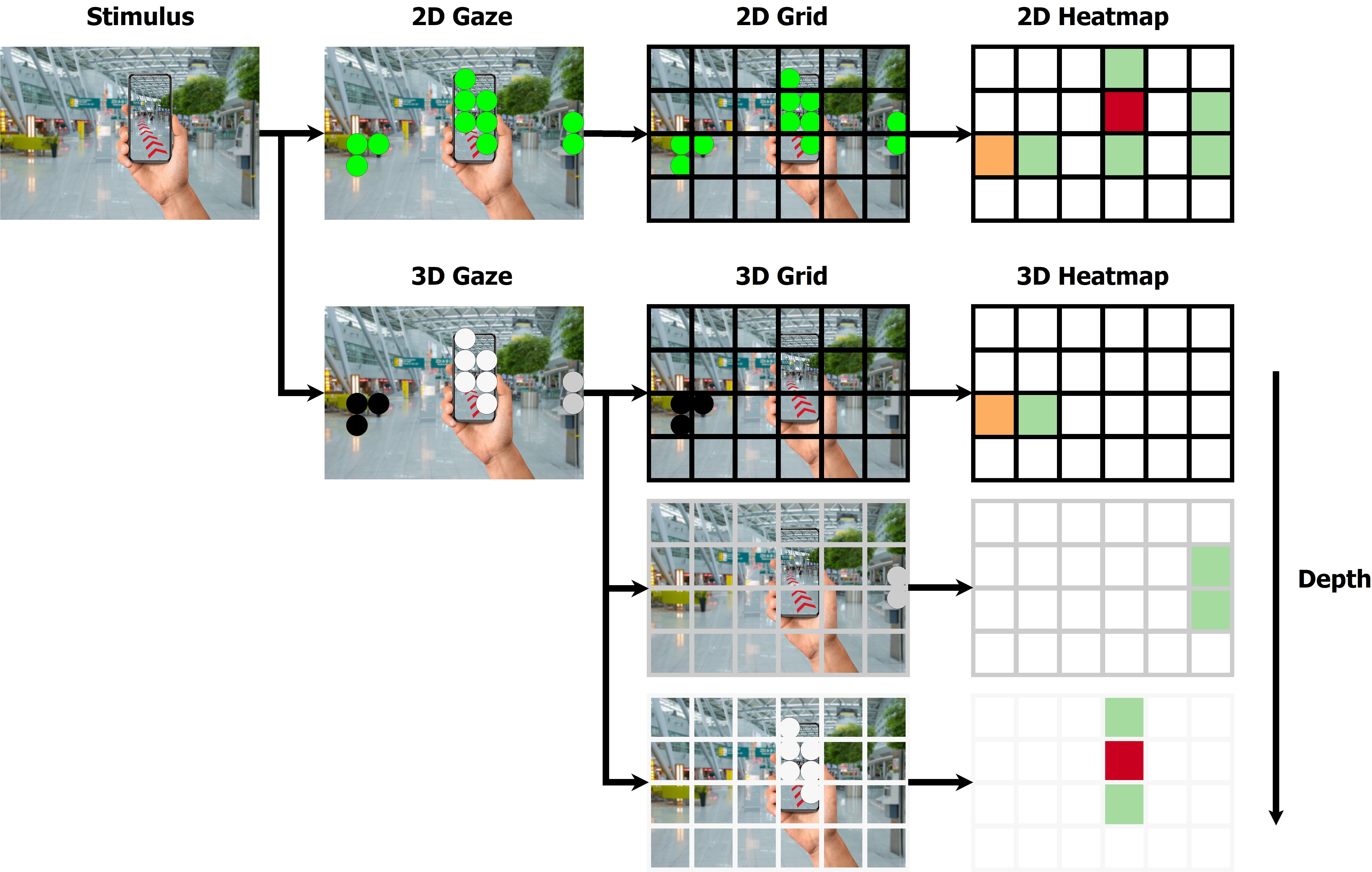}
	\caption{Creation of a 2D or 3D heatmap based on the gaze information and the stimulus resolution.}
	\label{fig:heatcreation}
\end{figure}

We decided to use a spatial distribution as a feature since this worked best in our initial evaluations. This spatial distribution is a heatmap as previously proposed by  \cite{fuhl2021gaze} to classify gaze position data. To create such a heatmap, the gaze position data of a temporal window are used, and the individual gaze positions are assigned to cells in the heatmap (grid).
Each time window results in one heatmap.
The assignment procedure is illustrated in \figref{fig:heatcreation}.
After the assignment, the heatmap is divided by the sum over all values to obtain a distribution. As an extension to the approach in \cite{fuhl2021gaze}, we extended the 2D heatmap to 3D. This was possible because the software used for gaze determination generates 3D gaze points~\cite{PISTOL} based on a $k$-nearest neighbor regression. In the case of the 3D heatmap, a cell is assigned to each gaze point based on its spatial position with the difference to the 2D heatmap that the depth or distance of the gaze points is additionally considered along the z-axis. An example of an assignment of gaze points to a 3D heatmap can be seen in \figref{fig:heatcreation}.

A formal description of the generation of the heatmap in 3D is given in Equation~\ref{eq:heat}.
\begin{equation}
    \operatorname{heat}\left(\, \round{\frac{p_x}{R_x} \cdot G_x},\, \round{\frac{p_y}{R_y} \cdot G_y},\, \round{\frac{p_z}{R_z} \cdot G_z}\, \right) \pluseq 1.
          \label{eq:heat}
\end{equation}
The gaze positions in x, y, and z coordinates in an Euclidean coordinate system are denoted by $p_x$,~$p_y$,~and~$p_z$, respectively. The constants $R_x$,~$R_y$,~and~$R_z$ represent the maximum resolution of the stimulus in x and y direction and the maximum depth supported by the software Pistol~\cite{PISTOL}. By dividing the gaze points by the maximum resolution, these ranges are normalized between~ $ 0 $ and~$ 1 $. Subsequently, these values are multiplied by the number of grid cells ($G_x$,~$G_y$,~and~$G_z$) and rounded to the nearest integers, denoted by~``$\round{\cdot}$''. These new values correspond to the index in the heatmap and the selected cell is incremented by one, denoted by~``$\pluseq$''. In the case of a 2D heatmap, the cell for depth (z~coordinate) is fixed at one.

Equation~\ref{eq:norm} describes the normalization of the heatmap in 3D and 2D since for the 2D case there would be only one depth.
\begin{equation}
    \operatorname{heat}(x,y,z) = \frac{\operatorname{heat}(x,y,z) }{\sum_{i=1}^{G_x} \sum_{j=1}^{G_y} \sum_{k=1}^{G_z} \operatorname{heat}(i,j,k)}.
          \label{eq:norm}
\end{equation}
Our normalization sums up the entire heatmap and divides each value of the heatmap by this sum. The variables~$x$,~$y$,~and~$z$ are the indexes to the heatmap corresponding to the x-axis, y-axis, and z-axis. As in Equation~\ref{eq:heat}, the variables $G_x$,~$G_y$,~and~$G_z$ are the maximum amount of grid cells in the heatmap.
Finally, the one-dimensional vector resulting from the flattening of the heat map can be used as an input feature for various machine learning techniques.

	\section{Study Design \& Data Acquisition}
In this section, we describe the dataset we used. In order to evaluate our approach, a  dataset was required which contains not only eye tracking information but also, in addition to the gaze points, the bounding boxes of the objects that the participants were looking at.
Since, to the best of our knowledge, no such dataset exists or is publicly available, we collected a novel data set.
In the beginning, we gave each of the five subjects an introduction to the recording procedure.
Each recording started with a calibration.
To this end, the participants were instructed to stand \unit{0.5}{\meter} in front of the calibration marker and look at its center. 
They were then asked to walk backward for about five meters, slowly circling their head while fixating on the center of the marker throughout.
Subsequently, the subjects were allowed to move freely around the site, both inside and outside the building, a university complex with several floors, corridors, and offices on the inside and a street, parking lots, and green areas on the outside.
In this course, they should look at arbitrary objects they encountered, such as first aid kits, fire extinguishers, light switches, door signs, street signs, trees, and bicycles, among others.
There was no specification as to how long they were supposed to look at the objects.
To evaluate gaze accuracy, the participants were asked to look at the calibration marker again at the end of each recording.
Initially, five meters away, they had to move towards the marker with head rotations until they were standing directly in front of it.
Due to the different distances, we are able to estimate depth information in both calibration and evaluation data via the tool Pistol~\cite{PISTOL}.
All recordings were conducted with the Pupil Invisible eye tracker, a head-mounted eye tracker developed by Pupil Labs \cite{PupilLabs}, whose scene camera provides RGB images with a resolution of 1088 $\times$ 1080.
Each participant captured three recordings (each recording was about five minutes long, including calibration and evaluation), resulting in 15 videos in total.
Only one recording could not be used due to an incorrectly performed calibration, where the marker was in the image of the scene for a while without the subject looking at it.
Even though the tool Pistol filters out a certain amount of erroneous data, the incorrect portion was too large during the calibration phase.
This led to a total length of about one hour of recording, consisting of $ 102\,620 $ frames of which $ 27\,946 $ contained objects.

Finally, we labeled the obtained data with DarkLabel \cite{DarkLabel}.
The objects in the dataset, respectively their bounding boxes, have the characteristics shown in \tblref{tbl:datasetStats}.
\begin{table}[htb]
    \caption{Size distribution of the objects in the dataset. The numbers indicate the size of the respective bounding box in pixels. Note that the columns do not have to originate from the same bounding box. The mean is denoted by $ \mu $ and the standard deviation by $ \sigma $.}
    \label{tbl:datasetStats}
    \centering
    \begin{tabular}{c|rrr}
        & Width & Height & Size \\
        \midrule
        min     & 16    & 18    & 288 \\
        max     & 1056  & 955   & 830\,484 \\
        $\mu$   & 177   & 227   & 61\,456 \\
        $\sigma$& 154   & 185   & 103\,384 \\
    \end{tabular}
\end{table}

\begin{figure}[htb]
	\centering
	\begin{subfigure}{.33\columnwidth}
		\centering
	    \includegraphics[width=.95\linewidth]{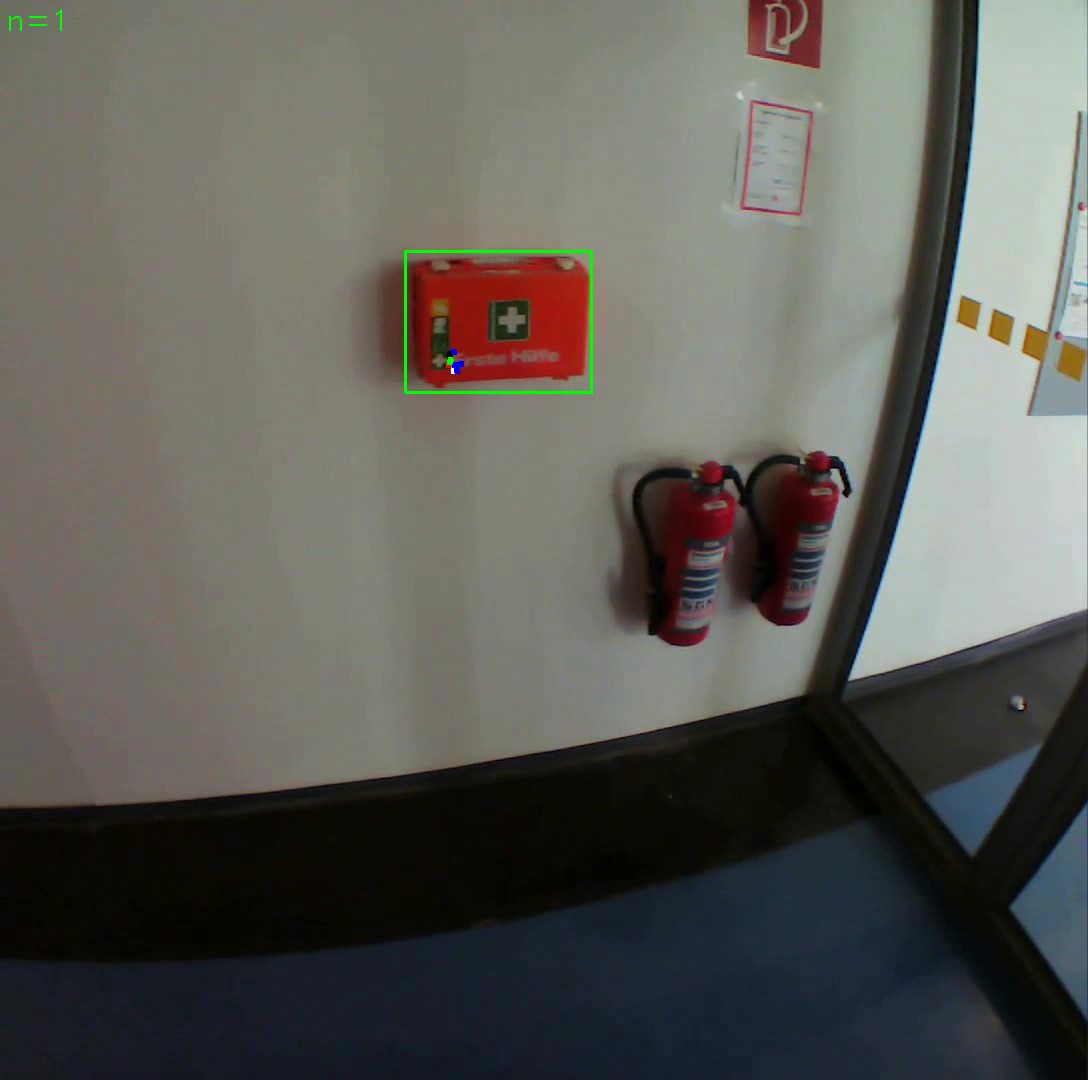}
		\caption{}
		\label{subfig:with1}
	\end{subfigure}%
	\begin{subfigure}{.33\columnwidth}
		\centering
	    \includegraphics[width=.95\linewidth]{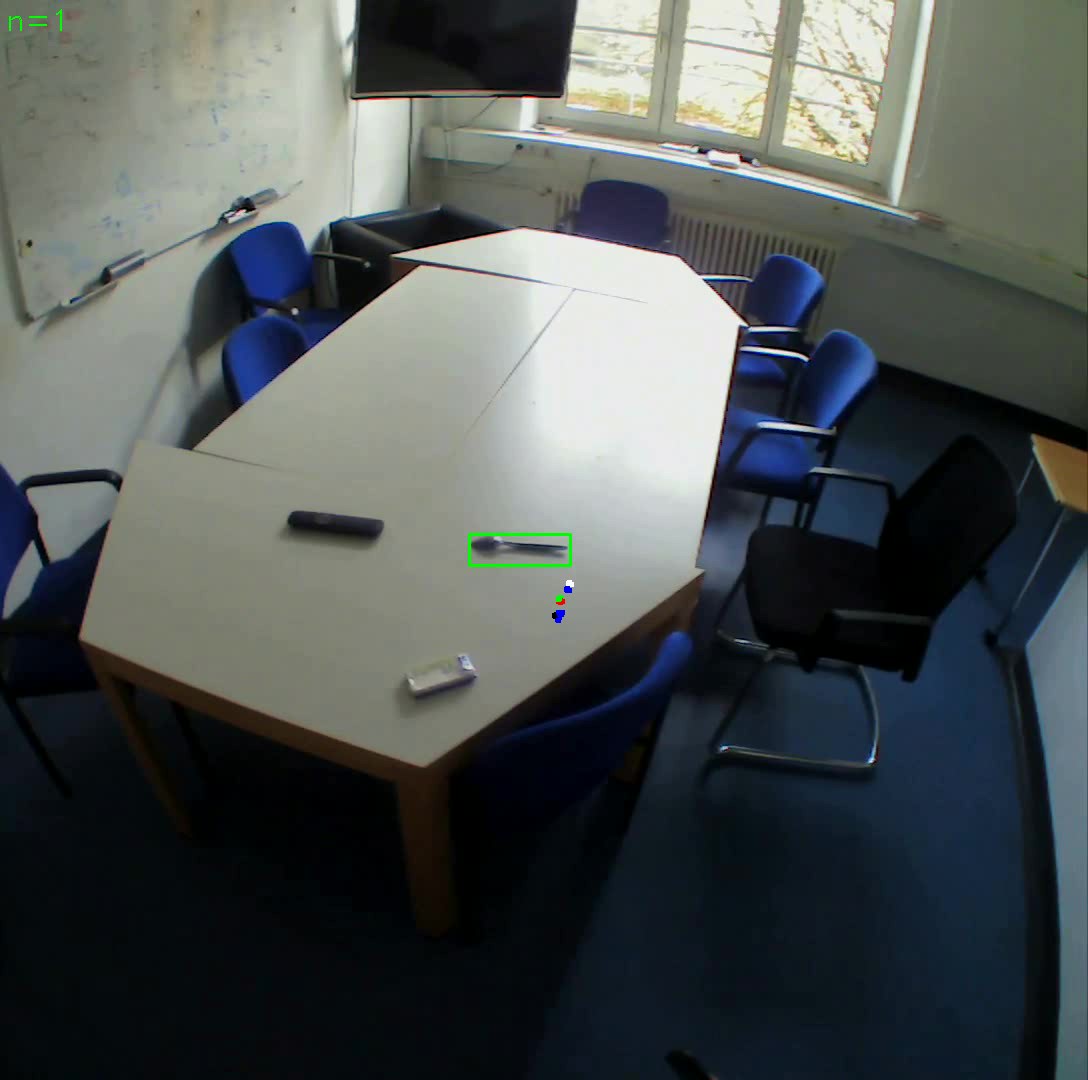}
		\caption{}
		\label{subfig:without1}
	\end{subfigure}%
	\begin{subfigure}{.33\columnwidth}
		\centering
	    \includegraphics[width=.95\linewidth]{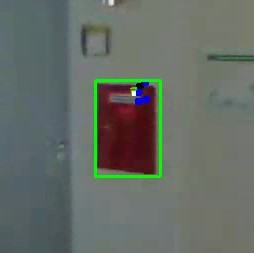}
		\caption{}
		\label{subfig:with2}
	\end{subfigure}
	\par \bigskip
	\begin{subfigure}{.33\columnwidth}
		\centering
	    \includegraphics[width=.95\linewidth]{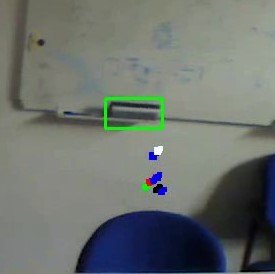}
		\caption{}
		\label{subfig:without2}
	\end{subfigure}%
	\begin{subfigure}{.33\columnwidth}
		\centering
	    \includegraphics[width=.95\linewidth]{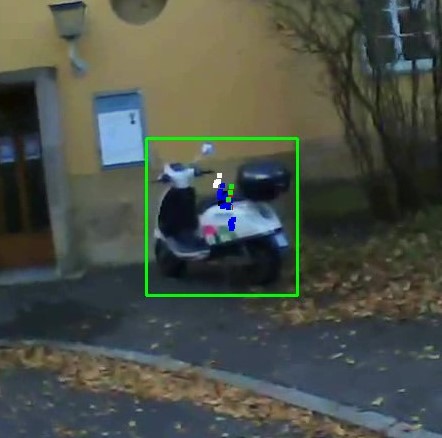}
		\caption{}
		\label{subfig:with3}
	\end{subfigure}%
	\begin{subfigure}{.33\columnwidth}
		\centering
	    \includegraphics[width=.95\linewidth]{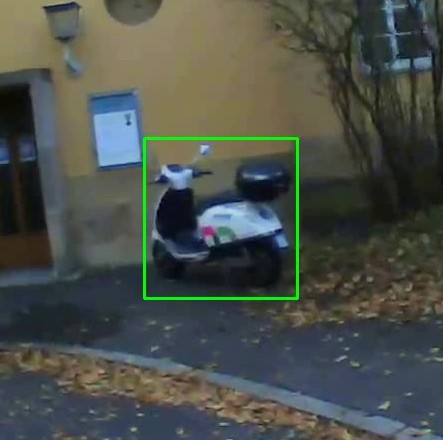}
		\caption{}
		\label{subfig:without3}
	\end{subfigure}%
	\caption{The images show exemplary moments of our data, where the objects that were consciously observed are labeled with a bounding box. The first two images \subfigref{subfig:with1} and \subfigref{subfig:without1} show the entire field of view of the scene camera (1088 $\times$ 1080), while the other images~\subfigref{subfig:with2}--\subfigref{subfig:without3} have been zoomed in to better show the bounding box and the various estimated gaze points. The gaze points are colored depending on the estimation method used by Pistol~\cite{PISTOL}. In \subfigref{subfig:without1} and \subfigref{subfig:without2}, the gaze points do not lie within the bounding boxes since the gaze estimation is --- especially for smaller objects --- not always completely accurate.
	As in the last image~\subfigref{subfig:without3}, there may also be occasionally no gaze points in between.
	In light of these challenges, gaze-based object detection is a non-trivial problem.
	}
	\label{fig:labeledBoxes}
\end{figure}

\figref{fig:labeledBoxes} shows individual example moments from the recordings.
Due to the errors related to the gaze estimation, the gaze points are not always on the labeled object, even though the participant was actually looking at it.
In fact, even for a human, it is not always easy to determine the target object, and sometimes only possible considering the context and the observation of an image sequence.
This demonstrates quite clearly the difficulties and challenges associated with this task.
Our final, publicly available dataset only contains the gaze information and bounding boxes, yet no stimuli-related information.

	\section{Evaluation}
In this section, we evaluate the classification of the gaze points with respect to the affiliation to an object, and we try to extract the position and the size of the object from those. 
To this end, we applied a variety of different, well-established machine learning methods and list here a selection comprising the best of them.
For easy reproducibility of our approach, we restrict ourselves to Matlab's standard parameters of the used machine learning methods. In the classification experiments, we always specify the mean accuracy of a 5-fold cross validation. For the regression experiments, the mean error as a percentage of the image resolution from a 5-fold cross validation is given. We evaluated different heatmap grid sizes as well as different time window sizes.
Furthermore, we investigate the runtime and memory requirements of our method and compare it to state-of-the-art object detectors.
We conducted our evaluations on a computer system with Windows 10 as the operating system, an AMD Ryzen 9 3950X 16-core processor with 3.50 GHz, and 64 GB DDR4 Ram. All machine learning methods were implemented on the Matlab version 2021b.

The assignment of classes (object or no object) to time windows was done based on the presence of an annotated object in the time window. This means that if there was an annotated object in the time window, the class was set to one. If this was not the case and there was no annotated object in the time window, the class was assigned zero. In the regression, only time windows with an existing annotated object were used. Here, the parameters of the annotated object closest to the central timestamp of the time window were chosen. This was assigned because, in most cases, our subjects moved while looking at an object. Thus, there are usually different positions and sizes of bounding boxes in a time window. To compensate for this, the most central object in time was always chosen.
The conversion of the time windows into seconds can be roughly calculated using the formula $ \text{Sec} = T/(6.66 \cdot 30)$, where $ T $ is the size of the respective time window in frames (eye camera). This is due to the recording rates of the scene camera (30 frames per second) and the recording rate of the eye camera (200 frames per second) of the Pupil Invisible~\cite{PupilLabs} eye tracker. The $ 6.66 $ is the number of eye frames per scene image. An alternative formulation is to divide the time window size $ T $ by the number of eye frames per second, which is 200.

\begin{figure}[htb]
	\centering
    \includegraphics[width=0.9\linewidth]{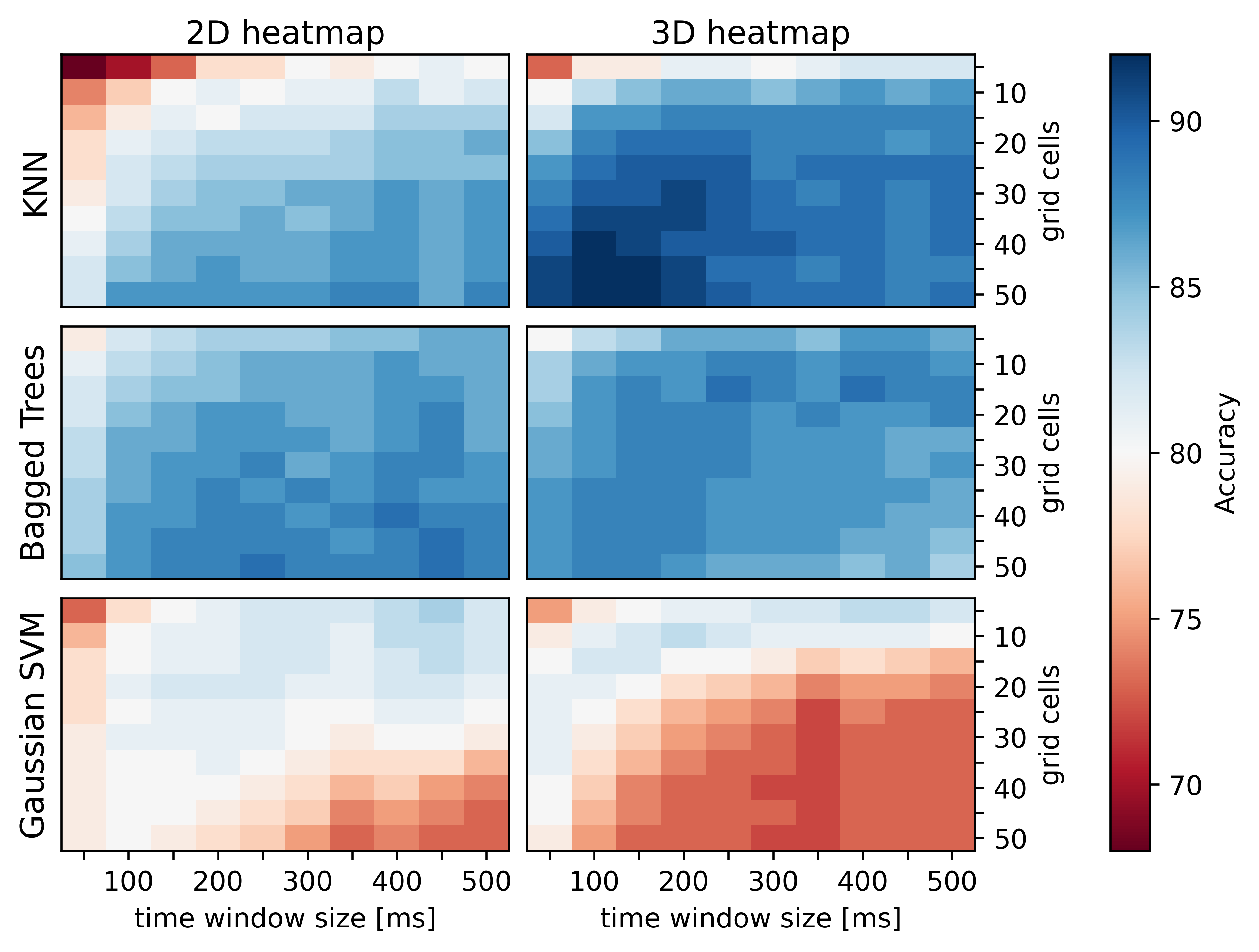}
    \caption{Classification results of the 2D and 3D heatmap features for different time window sizes (in ms), number of grid cells, and machine learning methods illustrated in a heatmap. The results are the average accuracy of a 5-fold cross validation.}
    \label{fig:evalClassification}
\end{figure}

\figref{fig:evalClassification} and \tblref{tbl:evalClassification} show the results of our classification experiment. 
\begin{table}[htb]
    \caption{Best and worst classification results of the 2D and 3D heatmap features. The mean is denoted by $ \mu $ and the standard deviation by $ \sigma $.}
    \label{tbl:evalClassification}
    \centering
    \begin{tabular}{cc|ccc}
        \multirow{2}{*}{Feature} & \multirow{2}{*}{ML} & \multicolumn{2}{c}{Accuracy} & \multirow{2}{*}{$\mu\pm\sigma$} \\
        & & Worst & Best & \\
        \midrule
        \multirow{3}{*}{2D heatmap} & KNN & 68 & 88 & $83.2\pm3.8$ \\
        & Bagged Trees & 79 & 89 & $86.3\pm1.9$ \\
        & Gaussian SVM & 73 & 84 & $79.4\pm2.6$ \\
        \midrule
        \multirow{3}{*}{3D heatmap} & KNN & 73 & 92 & $87.8\pm3.3$ \\
        & Bagged Trees & 80 & 89 & $86.8\pm1.3$ \\
        & Gaussian SVM & 72 & 83 & $76.5\pm3.6$ \\
    \end{tabular}
\end{table}
Comparing the results of the three methods, i.e., KNN, bagged trees, and Gaussian SVM, it can be seen that for the 2D heatmap feature, the approach based on bagged trees achieves the best results. Looking at the progression over the grid size and the time window size, we can see that the KNN and the bagged trees perform best with a high number of grid cells and large time windows. In the case of the Gaussian SVM, this is different, as this method performs best at a small number of grid cells but still large time windows. Moving on to the 3D heatmaps, the accuracy of the KNN method improves by 4 percent to 92 percent, which is also significantly better than the bagged trees. The bagged trees do not improve overall and remain at 89 percent.

The results of our regression experiment are shown in \tblref{tbl:evalRegressionDetailed}.
\begin{table*}[htb!]
	\caption{Regression error results as the average absolute error ($\cdot 10^{2}$) of a 5-fold cross validation, normalized to the image resolution. The columns X and Y denote the position of the bounding box, W is the width, and H is the height of the bounding box. The best values per method are highlighted in bold.}
	\label{tbl:evalRegressionDetailed}
	\centering
	\def\arraystretch{0.69}
    \setlength\tabcolsep{0.7pt}
	\begin{tabular}{ccc|cccc|cccc|cccc|cccc|cccc}
		\toprule
		&  &  & \multicolumn{20}{c}{Time window size (in ms)}\\
		Feat. & ML & Grid & \multicolumn{4}{c}{100} & \multicolumn{4}{c}{200} & \multicolumn{4}{c}{300} & \multicolumn{4}{c}{400} & \multicolumn{4}{c}{500}\\
		& & & X & Y & W & H & X & Y & W & H & X & Y & W & H & X & Y & W & H & X & Y & W & H\\
		\midrule
		\multirow{30}{*}{\rotatebox{90}{2D heatmap}} & \multirow{10}{*}{\rotatebox{90}{Gaussian Process}} & 5 & 7.0 & 8.0 & 13.8 & 16.4 & 7.2 & 7.9 & 13.4 & 16.1 & 7.7 & 8.2 & 13.9 & 16.1 & 7.8 & 8.0 & 13.7 & 16.1 & 8.8 & 7.3 & 13.4 & 16.3 \\
		& & 10 & \textbf{6.1} & \textbf{6.8} & 13.0 & 15.3 & 6.5 & 6.9 & 12.5 & 15.2 & 8.8 & 7.5 & 13.8 & 15.3 & 8.1 & 7.1 & 13.2 & 16.5 & 8.2 & 6.9 & 13.8 & 15.5 \\
		& & 15 & \textbf{6.1} & \textbf{6.8} & 12.7 & 15.3 & 7.8 & 9.5 & 13.5 & 16.8 & 9.4 & 8.4 & 14.3 & 15.7 & 8.3 & 7.7 & 12.4 & 15.8 & 8.4 & 7.1 & 13.7 & 15.8 \\
		& & 20 & 6.2 & 6.9 & 12.9 & 15.3 & 8.8 & 10.2 & 14.1 & 17.0 & 9.4 & 9.9 & 13.7 & 16.3 & 8.1 & 7.6 & 12.4 & 15.4 & 8.5 & 8.5 & 13.2 & 17.5 \\
		& & 25 & 6.8 & \textbf{6.8} & 12.7 & 15.2 & 8.8 & 8.8 & 13.9 & 16.8 & 9.3 & 9.2 & 14.3 & 15.9 & 8.4 & 8.9 & 13.0 & 15.8 & 8.5 & 8.5 & 13.4 & 17.0 \\
		& & 30 & 8.6 & \textbf{6.8} & 12.9 & 15.4 & 8.8 & 10.8 & 14.0 & 17.3 & 9.3 & 11.1 & 14.3 & 17.7 & 8.7 & 10.6 & 14.1 & 17.3 & 8.9 & 8.6 & 13.7 & 17.5 \\
		& & 35 & 7.9 & \textbf{6.8} & 12.7 & \textbf{15.1} & 8.8 & 10.8 & 13.4 & 17.3 & 9.3 & 11.1 & 14.3 & 17.7 & 8.7 & 10.6 & 14.1 & 17.3 & 8.6 & 8.6 & 13.9 & 17.5 \\
		& & 40 & 8.1 & \textbf{6.8} & \textbf{12.2} & 16.4 & 9.1 & 10.8 & 12.7 & 17.3 & 9.3 & 11.1 & 14.3 & 17.7 & 8.7 & 10.6 & 14.1 & 17.3 & 8.6 & 10.3 & 13.8 & 17.5 \\
		& & 45 & 8.1 & \textbf{6.8} & 12.6 & 16.9 & 9.1 & 10.8 & 14.2 & 17.3 & 9.3 & 11.1 & 14.3 & 17.7 & 8.7 & 10.6 & 14.1 & 17.3 & 9.1 & 10.3 & 14.4 & 17.5 \\
		& & 50 & 8.0 & \textbf{6.8} & 12.6 & 16.8 & 9.1 & 10.8 & 14.2 & 17.3 & 9.3 & 11.1 & 14.3 & 17.7 & 8.7 & 10.6 & 14.1 & 17.3 & 9.1 & 10.3 & 14.4 & 17.5 \\ \cmidrule(lr{.5em}){2-23}
		& \multirow{10}{*}{\rotatebox{90}{Bagged Trees}} & 5 & 7.0 & 7.9 & 13.5 & 16.4 & 7.2 & 7.7 & 13.2 & 16.3 & 7.6 & 8.1 & 13.4 & 16.1 & 7.4 & 7.8 & 13.5 & 15.7 & 7.8 & 7.3 & 13.6 & 15.7 \\
		& & 10 & \textbf{6.4} & 7.0 & 13.0 & 15.3 & 6.7 & 7.1 & 12.6 & 15.2 & 7.4 & 7.9 & 12.5 & 15.3 & 7.0 & 7.7 & 12.5 & 14.9 & 7.4 & 7.3 & 12.5 & 15.0\\
		& & 15 & \textbf{6.4} & 7.1 & 12.6 & 15.1 & 6.8 & 7.3 & 12.3 & 14.9 & 7.5 & 8.0 & 12.5 & 15.3 & 7.2 & 7.9 & \textbf{12.0} & 15.0 & 7.8 & 7.2 & 12.2 & 15.1 \\
		& & 20 & \textbf{6.4} & \textbf{6.9}& 12.7 & 14.8 & 6.9 & 7.1 & 12.2 & 14.8 & 7.6 & 8.0 & 12.2 & 15.4 & 7.1 & 7.8 & 12.2 & 15.0 & 7.7 & 7.2 & 12.1 & 14.9 \\
		& & 25 & 6.5 & 7.0 & 12.6 & 14.8 & 7.1 & 7.4 & 12.2 & 14.7 & 7.7 & 8.2 & 12.1 & 14.8 & 7.2 & 7.9 & 12.1 & 14.6 & 7.7 & 7.4 & \textbf{12.0}  & 14.7 \\
		& & 30 & 6.5 & \textbf{6.9} & 12.4 & 14.7 & 6.9 & 7.3 & 12.2 & 14.6 & 7.6 & 8.2 & 12.4 & 15.0 & 7.3 & 7.9 & 12.2 & 14.7 & 7.7 & 7.4 & 12.3 & 15.1 \\
		& & 35 & \textbf{6.4} & 7.0 & 12.2 & 14.5 & 7.1 & 7.3 & 12.1 & 14.4 & 7.6 & 8.2 & 12.1 & 14.8 & 7.3 & 7.9 & 12.1 & 14.4 & 7.8 & 7.5 & 12.1 & 14.7 \\
		& & 40 & 6.5 & 7.0 & 12.1 & \textbf{14.3}  & 7.1 & 7.3 & \textbf{12.0}  & 14.4 & 7.6 & 8.2 & \textbf{12.0}  & 14.5 & 7.4 & 8.0 & 12.1 & 14.6 & 7.8 & 7.4 & \textbf{12.0} & 14.4 \\
		& & 45 & 6.5 & 7.0 & 12.2 & 14.4 & 7.1 & 7.4 & 12.1 & 14.5 & 7.6 & 8.3 & 12.1 & 14.8 & 7.2 & 8.0 & \textbf{12.0}  & 14.5 & 7.6 & 7.6 & \textbf{12.0}  & 14.8 \\
		& & 50 & 6.6 & 7.1 & 12.3 & 14.4 & 7.1 & 7.4 & 12.2 & 14.4 & 7.6 & 8.3 & 12.3 & 14.5 & 7.3 & 8.1 & 12.1 & 14.6 & 7.8 & 7.5 & \textbf{12.0}  & 15.0 \\ \cmidrule(lr{.5em}){2-23}
		& \multirow{10}{*}{\rotatebox{90}{Gaussian SVM}} & 5 & 7.2 & 8.3 & 14.6 & 17.2 & 7.4 & 8.2 & 14.6 & 16.8 & 8.1 & 8.7 & 14.6 & 17.3 & 7.6 & 8.3 & 14.3 & 16.5 & 7.7 & 7.9 & 14.7 & 16.8\\
		& & 10 & 6.5 & 7.1 & 14.5 & 16.9 & 6.8 & 7.0 & 14.3 & 16.6 & 7.5 & 7.7 & 14.4 & 17.2 & 7.0 & 7.3 & 14.0 & 16.4 & 7.3 & 6.9 & 14.6 & 16.7\\
		& & 15 & \textbf{6.4}  & 7.0 & 14.4 & 16.6 & 6.9 & 7.0 & 13.9 & 16.2 & 7.6 & 7.8 & 14.0 & 16.7 & 7.1 & 7.4 & 13.8 & 16.0 & 7.5 & 7.1 & 14.2 & 16.1 \\
		& & 20 & 6.5 & 7.0 & 14.4 & 16.6 & 6.9 & 7.2 & 14.1 & 16.4 & 7.5 & 7.9 & 14.2 & 16.8 & 7.1 & 7.3 & 13.9 & 16.1 & 7.6 & 7.1 & 14.3 & 16.3 \\
		& & 25 & 6.5 & \textbf{6.9} & 14.0 & 16.4 & 7.0 & 7.2 & 13.8 & 16.0 & 7.6 & 7.8 & 13.9 & 16.5 & 7.2 & 7.6 & 13.8 & 15.8 & 7.7 & 7.2 & 14.1 & 15.9 \\
		& & 30 & 6.5 & \textbf{6.9} & 14.1 & 16.4 & 7.0 & 7.2 & 14.0 & 16.4 & 7.7 & 8.1 & 14.1 & 16.8 & 7.3 & 7.8 & 13.7 & 16.0 & 7.8 & 7.5 & 14.1 & 16.1 \\
		& & 35 & \textbf{6.4} & 7.0 & 13.7 & 16.0 & 7.0 & 7.2 & 13.7 & 15.8 & 7.7 & 8.0 & 13.7 & 16.1 & 7.3 & 7.8 & \textbf{13.4} & 15.6 & 7.7 & 7.4 & 13.7 & 15.7 \\
		& & 40 & 6.5 & 7.0 & 13.7 & 16.0 & 7.1 & 7.5 & 13.7 & 15.9 & 7.8 & 8.1 & 13.8 & 16.2 & 7.4 & 7.8 & 13.5 & \textbf{15.5} & 7.8 & 7.7 & 13.8 & 15.7 \\
		& & 45 & 6.6 & 7.0 & 13.9 & 16.3 & 7.1 & 7.4 & 13.8 & 16.0 & 7.8 & 8.3 & 14.0 & 16.6 & 7.5 & 8.0 & 13.7 & 16.0 & 7.9 & 7.7 & 14.1 & 16.2 \\
		& & 50 & 6.5 & 7.0 & 13.9 & 16.1 & 7.1 & 7.6 & 13.9 & 16.0 & 7.9 & 8.3 & 14.1 & 16.4 & 7.5 & 8.1 & 13.9 & 15.8 & 7.9 & 7.9 & 14.2 & 16.0 \\ \midrule
		\multirow{30}{*}{\rotatebox{90}{3D heatmap}} & \multirow{10}{*}{\rotatebox{90}{Gaussian Process}} & 5 & 6.7 & 7.5 & 12.8 & 15.4 & 7.1 & 7.3 & 12.4 & 15.7 & 9.4 & 8.6 & 13.1 & 16.6 & 8.6 & 8.4 & 13.2 & 15.6 & 8.9 & 7.3 & 13.8 & 15.2 \\
		& & 10 & \textbf{5.8} & 6.2 & 11.4 & 12.9 & 7.6 & 7.6 & 12.2 & 16.8 & 9.3 & 11.1 & 14.3 & 17.7 & 8.7 & 10.6 & 14.1 & 17.3 & 8.9 & 8.6 & 12.9 & 17.5 \\
		& & 15 & 6.5 & \textbf{6.0} & 10.4 & \textbf{11.6} & 9.1 & 10.8 & 14.2 & 17.2 & 9.3 & 11.1 & 14.3 & 17.7 & 8.7 & 10.6 & 14.1 & 17.3 & 9.1 & 10.3 & 14.4 & 17.5 \\
		& & 20 & 7.4 & 8.3 & \textbf{9.9} & 16.3 & 9.1 & 10.8 & 14.2 & 17.3 & 9.3 & 11.1 & 14.3 & 17.7 & 8.7 & 10.6 & 14.1 & 17.3 & 9.1 & 10.3 & 14.4 & 17.5 \\
		& & 25 & 9.1 & 10.8 & 14.3 & 17.4 & 9.1 & 10.8 & 14.2 & 17.3 & 9.3 & 11.1 & 14.3 & 17.7 & 8.7 & 10.6 & 14.1 & 17.3 & 9.1 & 10.3 & 14.4 & 17.5 \\
		& & 30 & 9.1 & 10.8 & 14.3 & 17.4 & 9.1 & 10.8 & 14.2 & 17.3 & 9.3 & 11.1 & 14.3 & 17.7 & 8.7 & 10.6 & 14.1 & 17.3 & 9.1 & 10.3 & 14.4 & 17.5 \\
		& & 35 & 9.1 & 10.8 & 14.3 & 17.4 & 9.1 & 10.8 & 14.2 & 17.3 & 9.3 & 11.1 & 14.3 & 17.7 & 8.7 & 10.6 & 14.1 & 17.3 & 9.1 & 10.3 & 14.4 & 17.5 \\
		& & 40 & 9.1 & 10.8 & 14.3 & 17.4 & 9.1 & 10.8 & 14.2 & 17.3 & 9.3 & 11.1 & 14.3 & 17.7 & 8.7 & 10.6 & 14.1 & 17.3 & 9.1 & 10.3 & 14.4 & 17.5 \\
		& & 45 & 9.1 & 10.8 & 14.3 & 17.4 & 9.1 & 10.8 & 14.2 & 17.3 & 9.3 & 11.1 & 14.3 & 17.7 & 8.7 & 10.6 & 14.1 & 17.3 & 9.1 & 10.3 & 14.4 & 17.5 \\
		& & 50 & 9.1 & 10.8 & 14.3 & 17.4 & 9.1 & 10.8 & 14.2 & 17.3 & 9.3 & 11.1 & 14.3 & 17.7 & 8.7 & 10.6 & 14.1 & 17.3 & 9.1 & 10.3 & 14.4 & 17.5 \\ \cmidrule(lr{.5em}){2-23}
		& \multirow{10}{*}{\rotatebox{90}{Bagged Trees}} & 5 & 6.7 & 7.4 & 12.7 & 15.2 & 7.1 & 7.5 & 12.3 & 15.0 & 7.9 & 7.9 & 12.5 & 15.2 & 7.6 & 7.7 & 12.5 & 15.0 & 7.8 & 7.5 & 12.6 & 15.2 \\
		& & 10 & \textbf{6.4} & \textbf{6.7} & 11.8 & 13.8 & 7.0 & 7.2 & 11.5 & 14.0 & 7.6 & 8.1 & 11.9 & 14.7 & 7.3 & 8.2 & 11.6 & 14.1 & 7.7 & 7.8 & 11.4 & 14.3\\
		& & 15 & \textbf{6.4} & 6.8 & 11.1 & 12.8 & 7.1 & 7.3 & 11.0 & 13.0 & 7.8 & 8.3 & 11.4 & 13.6 & 7.3 & 8.3 & 11.6 & 13.3 & 8.2 & 7.7 & 11.1 & 13.4 \\
		& & 20 & 6.7 & 7.0 & 11.0 & 12.7 & 7.2 & 7.3 & 11.0 & 13.1 & 7.9 & 8.2 & 11.1 & 13.4 & 7.5 & 8.2 & 11.2 & 13.5 & 8.1 & 7.8 & 10.9 & 13.6 \\
		& & 25 & 6.6 & 7.1 & 10.9 & 12.6 & 7.2 & 7.6 & 11.0 & 12.6 & 7.8 & 8.6 & 10.9 & 13.2 & 7.5 & 8.4 & 11.1 & 13.1 & 8.2 & 7.9 & 10.5 & 12.8\\
		& & 30 & 6.8 & 7.2 & 10.8 & \textbf{12.3} & 7.5 & 7.8 & 11.0 & 12.7 & 8.1 & 8.6 & 10.9 & 13.4 & 7.8 & 8.7 & 11.1 & 13.3 & 8.4 & 8.2 & 10.8 & 12.9 \\
		& & 35 & 7.0 & 7.3 & 11.0 & 13.0 & 7.4 & 7.8 & 11.2 & 13.1 & 8.1 & 8.7 & 11.0 & 13.8 & 7.7 & 8.6 & 11.0 & 13.8 & 8.3 & 8.2 & 10.8 & 13.3 \\
		& & 40 & 7.1 & 7.5 & 10.7 & 12.6 & 7.5 & 8.0 & 10.9 & 12.8 & 8.1 & 8.8 & 11.0 & 13.5 & 7.7 & 8.8 & 10.6 & 13.4 & 8.5 & 8.3 & \textbf{10.5} & 13.3 \\
		& & 45 & 7.2 & 7.6 & 10.7 & 12.6 & 7.6 & 7.9 & 11.0 & 13.1 & 8.3 & 8.8 & 11.1 & 13.7 & 7.8 & 8.9 & 11.0 & 13.4 & 8.5 & 8.5 & 10.9 & 13.3 \\
		& & 50 & 7.2 & 7.7 & 10.9 & 12.9 & 7.6 & 8.1 & 11.1 & 13.2 & 8.2 & 9.0 & 11.0 & 13.6 & 7.8 & 9.0 & 11.0 & 13.5 & 8.3 & 8.7 & 10.9 & 13.9 \\ \cmidrule(lr{.5em}){2-23}
		& \multirow{10}{*}{\rotatebox{90}{Gaussian SVM}} & 5 & 7.1 & 7.9 & 14.0 & 16.9 & 7.3 & 7.8 & 13.9 & 16.7 & 8.0 & 8.3 & 13.7 & 17.1 & 7.6 & 8.1 & 13.4 & 16.4 & 8.0 & 7.6 & 13.2 & 16.3 \\
		& & 10 & 6.4 & 6.8 & 13.3 & 15.8 & 6.9 & 6.9 & 13.3 & 15.7 & 7.4 & 7.7 & 12.9 & 15.7 & 7.3 & 7.9 & 12.9 & 15.4 & 7.5 & 7.5 & 12.9 & 15.3 \\
		& & 15 & \textbf{6.2} & 6.5 & 12.4 & 14.1 & 7.0 & 6.8 & 12.6 & 14.4 & 7.5 & 7.8 & 12.7 & 15.0 & 7.3 & 8.0 & 12.8 & 14.7 & 7.7 & 7.8 & 13.0 & 14.8 \\
		& & 20 & \textbf{6.2} & 6.3 & 11.9 & 13.9 & 6.9 & 7.0 & 12.0 & 14.1 & 7.5 & 8.1 & 12.5 & 15.0 & 7.6 & 8.1 & 12.5 & 14.7 & 7.8 & 8.1 & 13.0 & 14.8 \\
		& & 25 & 6.3 & \textbf{6.2} & 11.5 & 13.5 & 6.9 & 7.2 & 11.8 & 13.8 & 7.7 & 8.4 & 12.3 & 14.8 & 7.5 & 8.4 & 12.4 & 14.4 & 8.0 & 8.3 & 13.0 & 15.0 \\
		& & 30 & 6.3 & 6.3 & 11.1 & 13.0 & 7.1 & 7.5 & 11.8 & 13.9 & 7.9 & 8.6 & 12.4 & 15.0 & 7.6 & 8.6 & 12.4 & 14.6 & 8.2 & 8.5 & 13.0 & 15.0 \\
		& & 35 & 6.4 & \textbf{6.2} & 11.2 & 13.0 & 7.3 & 7.5 & 11.8 & 13.9 & 8.1 & 8.7 & 12.4 & 15.1 & 7.7 & 8.6 & 12.4 & 14.8 & 8.2 & 8.6 & 13.2 & 15.5 \\
		& & 40 & 6.5 & 6.3 & \textbf{11.0} & \textbf{12.9} & 7.4 & 7.7 & 11.9 & 14.0 & 8.1 & 8.8 & 12.5 & 15.2 & 7.8 & 8.7 & 12.4 & 14.8 & 8.3 & 8.7 & 13.3 & 15.4 \\
		& & 45 & 6.6 & 6.6 & 11.3 & 13.1 & 7.5 & 7.8 & 12.1 & 14.1 & 8.3 & 8.9 & 12.6 & 15.3 & 7.9 & 8.8 & 12.7 & 15.0 & 8.4 & 8.8 & 13.5 & 15.7 \\
		& & 50 & 6.6 & 6.7 & 11.2 & 13.0 & 7.5 & 8.0 & 12.2 & 14.3 & 8.4 & 9.1 & 12.8 & 15.5 & 7.9 & 9.0 & 12.8 & 15.2 & 8.4 & 8.9 & 13.7 & 15.8 \\ \bottomrule
	\end{tabular}
\end{table*}
To calculate a reasonable pixel value from the results, one must multiply them by ten. This is because the results are entered relative to the screen resolution (values between zero and one) and the mean absolute error was multiplied by 100 to obtain the percentage. Since the scene resolution is approximately $1000 \times 1000$ pixels, multiplying by ten is sufficient.
Looking at the individual methods (Gaussian process regression, bagged trees, and Gaussian SVM), we see that all methods perform similarly well.
As expected, based on the spatial heatmap feature, the position estimation is the most accurate. In contrast, the regression of the bounding box size, using only gaze data an no stimuli, is even more difficult than the position estimation and therefore less accurate.
Comparing the results for the 2D and the 3D heatmap feature, the position results remain about the same, with some overall improvement.
In terms of bounding box size, the best results improve significantly for all of the three methods.
All in all, the Gaussian process method combined with the 3D heatmap feature performs best.

\begin{figure}[htb]
	\centering
	\includegraphics[width=0.49\linewidth]{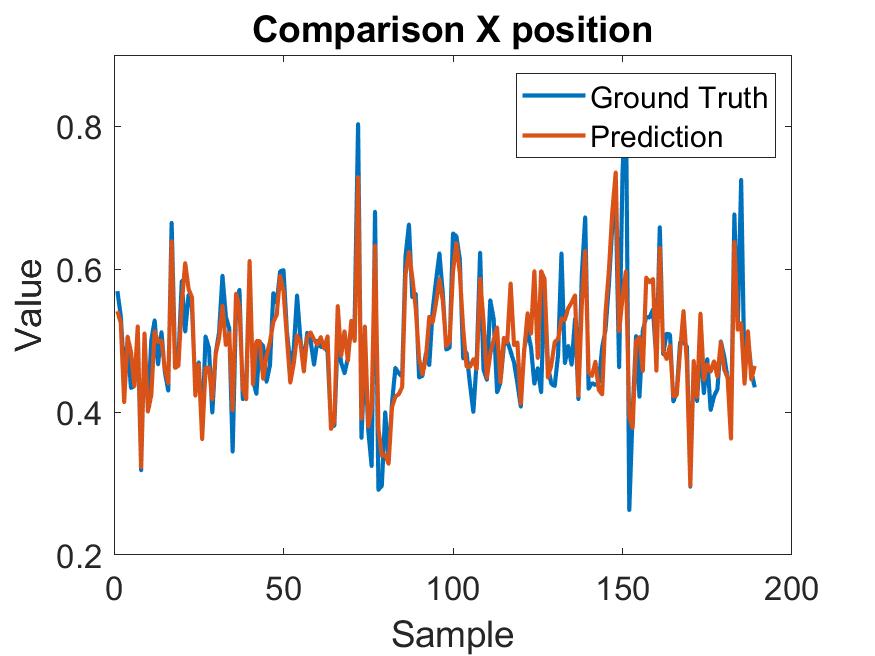}
	\includegraphics[width=0.49\linewidth]{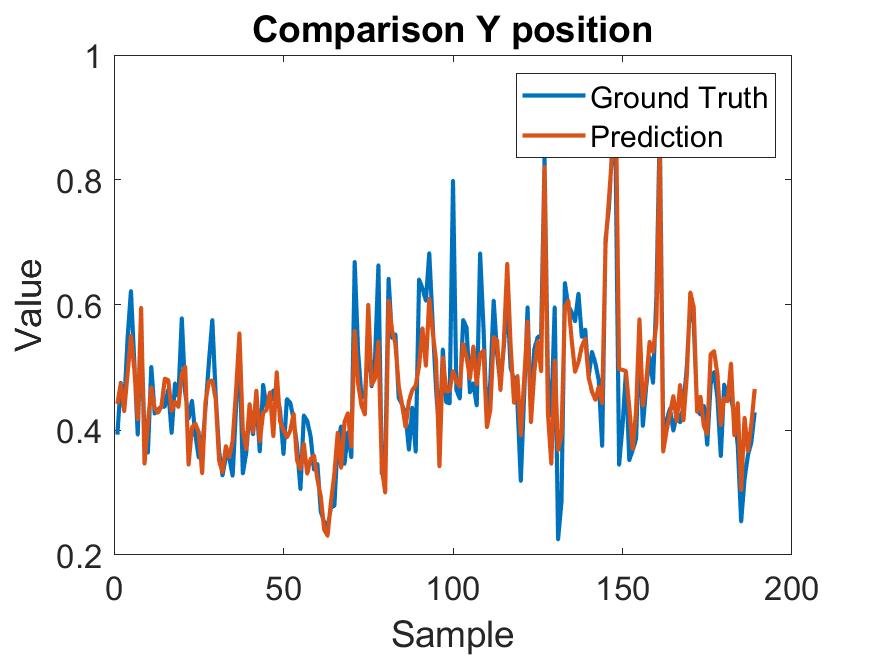}
	\par\bigskip
	\includegraphics[width=0.49\linewidth]{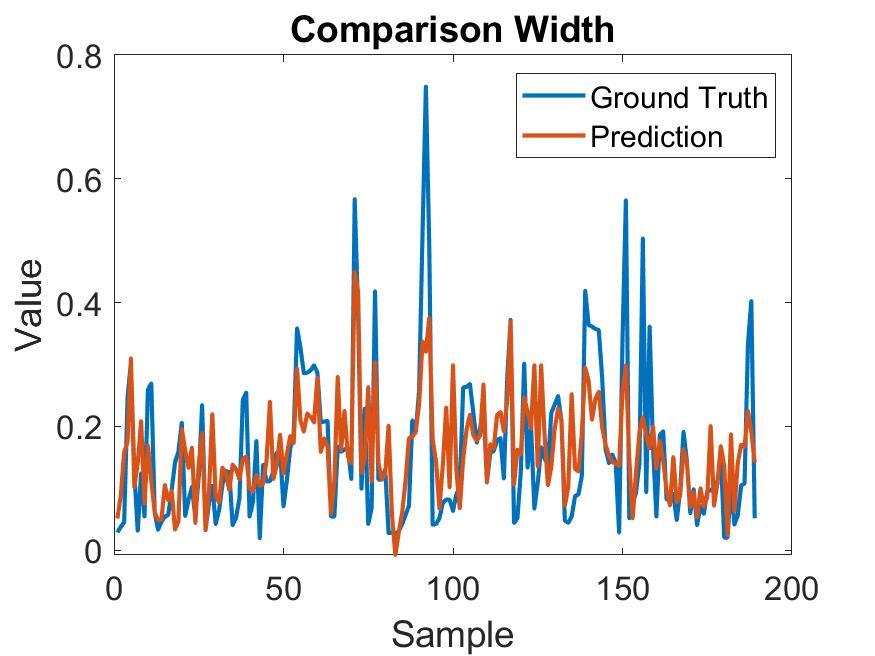}
	\includegraphics[width=0.49\linewidth]{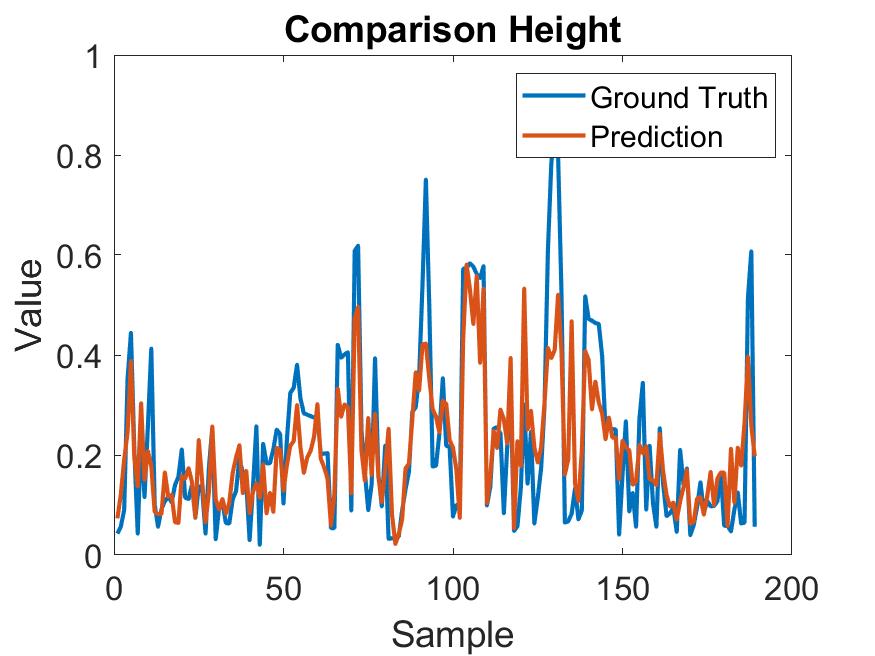}
	\caption{Qualitative evaluation of the bounding box parameter regression. Ground truth in blue and the prediction in red. The y-axis represents the value and the x-axis the sample number. The results are from the Gaussian Process Regression with a time window size of 100, a grid cell number of 15 and the 3D heatmap feature.}
	\label{fig:QualEval}
\end{figure}

\figref{fig:QualEval} shows a small part of the results of the Gaussian process regression in comparison to the ground truth. As can be seen, the determination of the X and Y position is quite accurate, whereas there are some outliers in the determination of the bounding box size. This is mainly due to the fact that in many cases the entire object was annotated, but not the whole object was observed. For this reason, the actual size cannot be estimated from the gaze data in these cases. Nevertheless, this is the way humans naturally look at objects in their environment, and it can be seen from the plots of the width and height of the bounding box that the method still works very well overall.

In the following, we will compare our approach with a baseline to better classify its performance.
It should be borne in mind that classical object detectors, such as FCOS~\cite{tian2019fcos} or Faster \mbox{R-CNN}~\cite{ren2015faster}, pursue a slightly different goal than we do.
Whereas in their case all objects are usually to be detected, we are primarily interested in the existence of an object of interest, that is, the one that the human is looking at.
Since classical object detectors only use images of the scene and do not obtain information about human gaze behavior, they cannot know whether a human has looked at an object and, if so, which object.
Thus it would be basically a matter of chance whether the statement that a person looks at an object is true or false.

With the regression task, the detection of all objects would be possible. Here, however, we encounter a different real-world problem, outside of laboratory conditions, which also makes our method so appealing.
Since we are in a wild world, the objects of interest are extremely diverse and their number tremendous.
The vast majority of objects in our dataset, such as doorknobs, light switches, and fire extinguishers, are simply not part of any publicly available data sets, such as Microsoft COCO~\cite{lin2014microsoft} or ImageNet~\cite{deng2009imagenet}, that are typically used for training.
Since the methods differ too much in this respect, we need a benchmark that covers more the commonalities.
Therefore, in the remainder of this section, we will establish a baseline comparison in terms of speed and computing resources.
As a baseline, we use state-of-the-art object detectors.
These include Faster \mbox{R-CNN}~\cite{ren2015faster}, FCOS~\cite{tian2019fcos}, and RetinaNet~\cite{lin2017focal}, each with a ResNet-50-FPN backbone~\cite{he2016deep},
SSDlite320~\cite{liu2016ssd} and Faster \mbox{R-CNN} both with a MobileNetV3 Large backbone~\cite{sandler2018mobilenetv2, howard2019searching}, as well as SSD300~\cite{liu2016ssd} with a VGG16 backbone~\cite{simonyan2014very}.
These are supplemented by the YOLOv5~\cite{jocher2022yolov5} variants YOLOv5n, YOLOv5s, YOLOv5m, YOLOv5l and YOLOv5x.
All of these object detectors were trained on Microsoft COCO and all backbones were trained on ImageNet.
In order to test the speed, we measured the runtime of all methods on the CPU for 1000 individual predictions, i.e. 1000 different inputs with a batch size of 1.
The resource consumption was determined by measuring the amount of memory required for a single input.
For our method with the heatmap input features, we used a time window size of 250\,ms.
For the classic object detectors, the $ 1088 \times 1080 \times 3 $ RGB images were used as input.
The results are shown in \tblref{tbl:evalResources}.
\begin{table*}[htb]
    \caption{Comparison of the required resources for the different input features. The time column indicates the execution time for 1000 different inputs at a batch size of one in seconds. The memory column specifies the required memory of a single input in kilobytes. For the 2D and 3D heatmap features, a time window size of 250\,ms was used.}
    \label{tbl:evalResources}
    \centering
    \setlength\tabcolsep{2pt}
    \begin{tabular}{cccccccccccc}
        Feature & ML & \multicolumn{5}{c}{Time [s]} & \multicolumn{5}{c}{Memory [KB]} \\
        \midrule
        & & \multicolumn{5}{c}{Grid cells} & \multicolumn{5}{c}{Grid cells} \\
        & & 10 & 20 & 30 & 40 & 50 & 10 & 20 & 30 & 40 & 50 \\
        \cmidrule(lr){3-7}
        \cmidrule(lr){8-12}
        \multirow{3}{*}{2D heatmap} & KNN & \pz2.2 & \pz4.8 & 10.8 & 18.4 & 27.0 & 116 & \pz259 & \pz424 & \pz607 & \pz810 \\
        & Bagged Trees & 52.1 & 56.6 & 57.8 & 58.9 & 60.0 & 954 & 1106 & 1134 & 1147 & 1164 \\
        & Gaussian SVM & \pz0.9 & \pz2.9 & \pz8.6 & 19.3 & 37.7 & 108 & \pz238 & \pz406 & \pz584 & \pz840 \\
        \midrule
        & & \multicolumn{5}{c}{Grid cells} & \multicolumn{5}{c}{Grid cells} \\
        & & 10 & 20 & 30 & 40 & 50 & 10 & 20 & 30 & 40 & 50 \\
        \cmidrule(lr){3-7}
        \cmidrule(lr){8-12}
        \multirow{3}{*}{3D heatmap} & KNN & 12.0 & \pz80.9 & 276.9 & \pz620.5 & 1137.2 & \pz328 & 1214 & 3045 & 6278 & 12769\\
        & Bagged Trees & 58.1 & \pz61.9 & \pz64.7 & \pz\pz66.9 & \pz\pz74.2 & 1226 & 1318 & 1467 & 1699 & \pz2020\\
        & Gaussian SVM & 11.9 & 154.9 & 610.6 & 1811.1 & 4061.6 & \pz325 & 1360 & 3650 & 7443 & 15368 \\
        \midrule
        \multirow{11}{*}{RGB Image} & Faster R-CNN~\cite{ren2015faster} (RN50) & \multicolumn{5}{c}{8\,705.3} & \multicolumn{5}{c}{1\,745\,456} \\
        & Faster R-CNN~\cite{ren2015faster} (MN) & \multicolumn{5}{c}{1\,205.6} & \multicolumn{5}{c}{\pz\,545\,400} \\
        & FCOS~\cite{tian2019fcos} & \multicolumn{5}{c}{4\,723.2} & \multicolumn{5}{c}{\pz\,995\,416} \\
        & RetinaNet~\cite{lin2017focal} & \multicolumn{5}{c}{5\,184.5} & \multicolumn{5}{c}{1\,390\,580} \\
        & SSD300~\cite{liu2016ssd} & \multicolumn{5}{c}{\pz900.8} & \multicolumn{5}{c}{\pz\,529\,744} \\
        & SSDlite320~\cite{liu2016ssd} & \multicolumn{5}{c}{\pz163.7} & \multicolumn{5}{c}{\pz\,293\,788} \\
        & YOLOv5n~\cite{jocher2022yolov5} & \multicolumn{5}{c}{\pz200.6} & \multicolumn{5}{c}{\pz\,270\,168} \\
        & YOLOv5s~\cite{jocher2022yolov5} & \multicolumn{5}{c}{\pz486.3} & \multicolumn{5}{c}{\pz\,312\,104} \\
        & YOLOv5m~\cite{jocher2022yolov5} & \multicolumn{5}{c}{1\,127.6} & \multicolumn{5}{c}{\pz\,421\,904} \\
        & YOLOv5l~\cite{jocher2022yolov5} & \multicolumn{5}{c}{2\,174.5} & \multicolumn{5}{c}{\pz\,622\,536} \\
        & YOLOv5x~\cite{jocher2022yolov5} & \multicolumn{5}{c}{3\,677.9} & \multicolumn{5}{c}{\pz\,940\,508} \\
    \end{tabular}
\end{table*}

The fastest is the Gaussian SVM with the 2D heatmap feature and a grid cell number of 10, taking less than 1 second for all 1000 predictions. The KNN is similarly fast with 2.2 seconds. Both methods increase with the number of grid cells as the grid cell size becomes smaller and thus the input features become larger.
The Bagged Trees are slower than KNN and Gaussian SVM for a small number of grid cells, but the runtime increases proportionally less as the number of grid cells increases.
Consequently, even with large input sizes, the runtime for the 3D heatmap feature is in the range of one minute for the 1000 predictions while the runtime for KNN and Gaussian SVM increases considerably from a few seconds to several minutes.
Nonetheless, it is immediately apparent that the runtime is in general significantly lower compared to the object detectors using the RGB images as input features.
While only the smaller models like YOLOv5n and SSDlite remain under three minutes, the other models are much slower. 
In particular, the computation time required by the popular Faster \mbox{R-CNN} with a \mbox{ResNet-50} backbone exceeds that of the Bagged Trees by a factor of over 100, even with the largest 3D heatmap input feature.

A similar picture emerges with respect to the RAM allocated for one single prediction.
The memory requirements of the bagged trees are larger at the beginning, but do not increase as much in proportion to the number of grid cells as for the KNN and the Gaussian SVM.
These even require less than 1\,MB of memory for small inputs and to some extent only a few 100\,KB.
The demand by the bagged trees is in the range of only one and two megabytes for all grid sizes.
Even in the case with the narrowest grid pattern.
Here, KNN and Gaussian SVM require over 12\,MB and 15\,MB, respectively.
However, this is substantially less than the most frugal neural network YOLOv5n, which needs around 270\,MB.
Faster \mbox{R-CNN} with the \mbox{ResNet-50} backbone requires the most memory with over 1.7\,GB.
Again, the factor is more than 100 times larger than for the Gaussian SVM with the maximum number of grid cells.
Compared to the Bagged Trees, it even exceeds 860 times.

In summary, our method is several orders of magnitude faster than conventional object detectors while requiring only a fraction of their resources.

	\section{Limitations}
We have shown that the gaze data can be used to detect objects and their bounding boxes. What is still missing for an actual application is the combinatorial use with, e.g., a robot, which learns to detect these objects. This application, of course, still brings some challenges like inaccurate bounding boxes or wrongly detected objects. However, with a robot or a machine learning the objects, there are more possibilities like depth data and the image information from different perspectives. Thus, there are also more possibilities for optimization in this application. Another limitation of our work is that we did not perform a parameter search for the machine learning methods. This means that our results are certainly far below the possible detection rates that can be achieved. However, our results are easy to reproduce and might be considered rather as a proof-of-concept, yet conveying the potential of gaze in such tasks.
	\section{Conclusion}
In this work, we addressed object detection based on gaze data as well as the regression of the bounding box parameters for the detected objects. Based on evaluations with different parameters of the size of the time window and the grid size of the heatmap feature, our results show that it is possible to detect objects and determine their bounding box based solely on gaze information.
This problem is not trivial since the subjects move, and thus, the parameters of the bounding box vary in the time window itself. 
Additionally, we have used a variety of machine learning methods to show that they work for solving such challenges.
Besides, the functionality of several machine learning methods proves that our heatmap feature, which we have extended to 3D, can be used efficiently for this problem.
In comparison to classical object detectors that use image input features, we have shown that object detection by means of our heatmap features is significantly faster while only requiring a fraction of the computational resources.
This is of major relevance due to the fact that robots usually have only limited computing capacity at their disposal and cannot be equipped with powerful graphics units as they consume a lot of power.

However, a significant amount of work remains for the future as we plan to extend our proof of concept to a real robot by making the gaze of the human collaborator accessible to it.
Our approach can serve as a foundation for future applications in the field of human-machine interaction and collaboration, where robots can learn new objects from humans through instant knowledge sharing.
Hence, we hope that our methods and our dataset can help to advance researchers in this challenging context.

	\bibliographystyle{plain}
	\bibliography{literature}
	
\end{document}